\documentclass[11pt]{article}

\usepackage[final]{acl}

\usepackage{times}
\usepackage{latexsym}

\usepackage{float}

\usepackage[T1]{fontenc}

\usepackage[utf8]{inputenc}

\usepackage{microtype}

\usepackage{inconsolata}

\usepackage{graphicx}

\usepackage{amsmath,amssymb,amsfonts}
\usepackage{algorithmic}
\usepackage{textcomp}
\usepackage{booktabs}
\usepackage{xcolor}
\usepackage{newfloat}
\usepackage{listings}
\usepackage{tabularx}
\usepackage{array}

\newcommand{\tool}{\textsc{Isee}}

\newcommand{\circled}[1]{{\large \textcircled{\footnotesize #1}}}

\title{
{\tool}: Interactive Semantic Enrichment for Database Fields
}

\author{
  \textbf{Yuan Tian\textsuperscript{1}},
  \textbf{Yiru Chen\textsuperscript{2}},
  \textbf{Rakesh R. Menon\textsuperscript{2}},
  \textbf{Zifan Liu\textsuperscript{2}},
  \textbf{Ting Cai\textsuperscript{3}},
  \\
  \textbf{Fei Wu\textsuperscript{2}},
  \textbf{Anudeep Chimakurthi\textsuperscript{2}},
  \textbf{Prashanthi Ramamurthy\textsuperscript{2}},
  \\
  \textbf{Sridevi Aishwariya Ganesan\textsuperscript{2}},
  \textbf{Kun Qian\textsuperscript{2}},
  \textbf{Yunyao Li\textsuperscript{2}}
\\
  \textsuperscript{1}Purdue University,
  \textsuperscript{2}Adobe,
  \textsuperscript{3}University of Wisconsin--- Madison 
}

\begin{document}
\maketitle

\begin{abstract}
LLM-based agents are increasingly being deployed for data-related tasks, including data sense-making, exploration, and retrieval.
However, their performance heavily depends on the clarity and completeness of data semantics. 
Unfortunately, in practice, many field descriptions remain ambiguous or incomplete, as much of the essential context (e.g., the meaning of a customized field) originates from users' domain knowledge and is rarely documented publicly. 
This gap restricts the agents' task performance in downstream tasks, such as entity-linking.
To bridge this gap, we introduce a novel and comprehensive \underline{\textbf{I}}nteractive \underline{\textbf{SE}}mantic \underline{\textbf{E}}nrichment system ({\tool}).
Given a data field description, {\tool} measures its quality through a scoring system, gathers domain knowledge, and collaboratively enriches the semantics with users. 
Through a user study, automated user simulation, quantitative evaluation, and case study, we demonstrate that {\tool} significantly reduces cognitive load, improves description quality, and enhances downstream task performance.
\end{abstract}

\section{Introduction}


\begin{figure}[ht]
\centering
\includegraphics[width=0.5\textwidth]{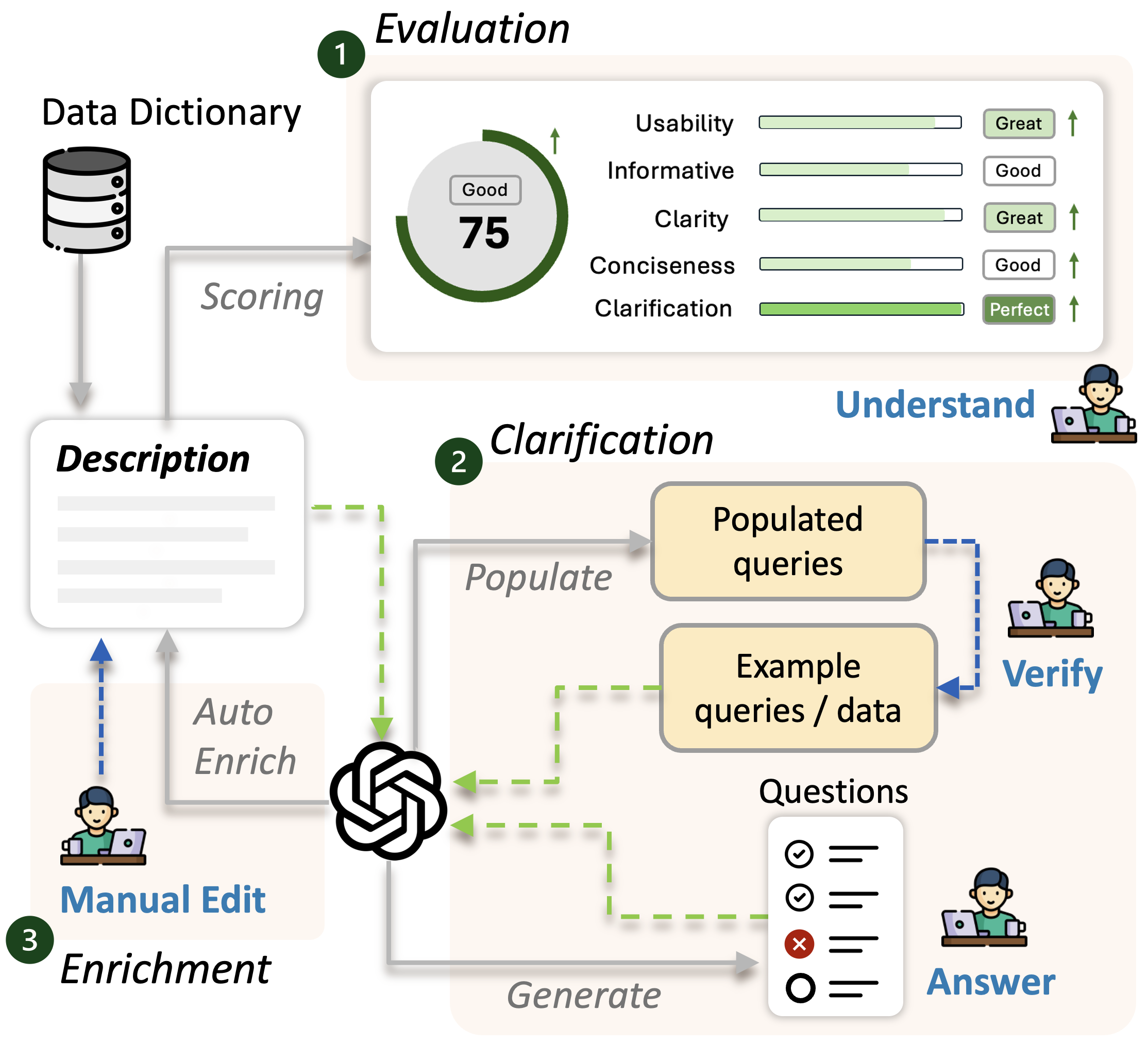}
\caption{Pipeline of {\tool}: (1) Users understand the quality of the current description by the scoring system. (2) Users provide feedback through clarification or query population. (3) Users refine the description. 
}
\label{fig:overview}
\end{figure}

LLM-based agents are increasingly deployed in enterprise tasks such as entity linking \cite{entity_linking1, entity_linking2} and natural language querying of databases \cite{empirical, steps, insight, pv_sql}. While academic environments assume well-documented or publicly available context, enterprise environments pose distinct challenges. Business databases often contain highly customized fields whose meaning is privately owned by customers.
Consequently, when field descriptions are ambiguous or incomplete, LLMs face challenges in interpreting their meaning, leading to errors in downstream tasks.
For example, when a user asks a natural language question, the entity-linking service is expected to accurately map the query to the corresponding mentioned fields. However, given a user-defined abbreviation name with unclear semantics, LLMs can only guess.
This is known as out-of-distribution~\cite{ood}, where LLMs fail to respond when the required knowledge is outside their parametric space.
In addition to training-based domain adaptation~\cite{sqlsynth, evo_schema, all_fem}, semantic enrichment \cite{semantic_enrichment1, semantic_enrichment2, semantic_enrichment3} can efficiently mitigate this issue by explicitly describing the missing semantics in documentation. 
However, they come with inherent limitations in practice.
Existing semantic enrichment methods assume that the enrichment can be achieved using the general knowledge embedded in LLMs, while many rely on domain-specific expertise from end-users.
Without corresponding context, it is infeasible for LLMs to automatically infer the missing information and enrich these fields. 
Thus, we argue that data field enrichment has to keep the domain experts in the loop and actively solicit domain knowledge.

We present {\tool}, a novel \underline{\textbf{I}}nteractive \underline{\textbf{SE}}mantic \underline{\textbf{E}}nrichment system that automatically evaluates the quality of existing descriptions, efficiently elicits user knowledge, and collaboratively enhances descriptions with domain experts. 
{\tool} introduces three key novel features. 
First, a \textit{downstream-aware scoring system} comprehensively evaluates the quality of field descriptions across five dimensions, including usability, informativeness, clarity, conciseness, and readability. 
We develop unique algorithms for each dimension to measure a distinct aspect of the description quality. 
Second, a \textit{conditional query population} component allows domain experts to enrich field descriptions by simply verifying LLM-populated natural queries related to this field. 
Third, a \textit{taxonomy-guided clarification generator} leverages our proposed clarification taxonomy to generate specific, easy-to-answer questions to capture domain experts' knowledge.

We evaluated {\tool} through a user study, an automated user simulation, a quantitative evaluation, and a case study. 
According to the user study, {\tool} improved description enrichment accuracy by 119\%, 32\%, and 92\% compared to manual editing, candidate selection, and ChatGPT, respectively, while significantly reducing cognitive load.
According to the quantitative evaluation, the enriched descriptions further benefit downstream tasks, improving entity linking by 22\% and text-to-SQL by 12\% over automatic enrichment.

Our contributions include:
\begin{itemize}
    \item A scoring system that evaluates the quality of a given description through five complementary aspects.
    \item A clarification question generator based on a newly proposed clarification taxonomy tailored for semantic enrichment.
    \item An interactive system that integrates scoring, clarification, and conditional query population into a unified workflow.
    \item A thorough evaluation including a user study, an automated user simulation, a quantitative evaluation, and a case study.
\end{itemize}

\section{Related Work}

\subsection{Semantic Enrichment}
Prior work has investigated automatically generating natural language descriptions for database schemas.
\citet{wretblad2024synthetic} generate column descriptions using a single LLM prompt and evaluate on text-to-SQL tasks.
Several efforts extend this to both table and column levels~\cite{zhang2025autoddg, tabmeta, singh2025leveragingretrievalaugmentedgenerative, gao2025automaticdatabasedescriptiongeneration}: \citet{zhang2025autoddg} generate table-level descriptions to support readability, \citet{tabmeta} employ an LLM-as-a-judge framework to select high-quality outputs, and \citet{singh2025leveragingretrievalaugmentedgenerative} leverage curated business glossaries for enrichment.

While these methods demonstrate the utility of schema descriptions, they rely on LLM general knowledge or static glossaries and generate descriptions in a single pass without incorporating domain expert feedback.
In contrast, {\tool} keeps the domain expert in the loop, iteratively eliciting and integrating user knowledge that cannot be inferred from public sources alone.

\subsection{Human-AI Collaboration for Data}
Human-AI collaborative systems combine human knowledge with AI capabilities to tackle tasks that neither can solve well alone~\cite{hai1,hai2}.
In data management, recent systems adopt this paradigm for tasks such as SQL query construction~\cite{sqlucid}, data exploration~\cite{empirical}, and generation steering~\cite{spa}, where users guide and validate AI-generated outputs.
A key mechanism in such systems is clarification: asking targeted questions to resolve ambiguity and elicit missing information.
Prior work on clarifying questions in information retrieval~\cite{10.1145/3366423.3380126, worker_agent, Aliannejadi_2019} and NL interfaces~\cite{taxonomy1, taxonomy2} has shown that well-structured questions significantly reduce misunderstanding.
{\tool} builds on these ideas by combining a clarification taxonomy with a multi-dimensional scoring system and query population, enabling a structured and iterative pipeline tailored for data field enrichment.

\section{Method}
\label{sec:method}

\subsection{System Overview}


Figure~\ref{fig:overview} presents the pipeline of {\tool}. 
Starting with field descriptions from a data dictionary (e.g., a relational database),
{\tool} includes three iterative stages—\emph{Evaluation}, \emph{Clarification}, and \emph{Enrichment}.
In the first stage (\emph{Evaluation}), {\tool} evaluates the quality of the current description. 
Users are provided with multiple evaluation scores, each measuring a distinct aspect of the description.
Additionally, natural language (NL) explanations accompany each score to help users better understand the quality and current state of the description.
This enables users to understand the situation and identify areas for improvement.
In the second stage (\emph{Clarification}), once users understand the current description, they can actively contribute related information (e.g., relevant NL queries or data records) using the query population feature. Alternatively, they can passively provide feedback by answering generated clarification questions.
This stage facilitates users in efficiently providing missing semantic information.
In the third stage, by incorporating user feedback, {\tool} collaboratively suggests an updated field description, which users can further refine as needed.

This iterative loop, including three stages, continues until the description achieves a satisfactory level of quality.
{\tool} follows the design of Human-AI collaborative systems such as \cite{sqlucid, sqlsynth}, where humans guide and validate the enrichment process while AI assists with automation and scalability.
We use OpenAI's GPT-4o \cite{gpt4o} as the base LLM and \texttt{text-embedding-3-small}~\footnote{https://platform.openai.com/docs/models/text-embedding-3-small} as the embedding model in {\tool}.
We discuss details of each component in the following sections.

\subsection{Scoring System}
\label{sec:scoring}
{\tool} provides a scoring system that quantitatively measures the given description's quality across five dimensions (0--100 for each).
Unlike approaches that delegate scoring to an LLM~\cite{llm_score1, llm_score2, llm_score3}, each metric uses a dedicated algorithm, ensuring greater stability than LLM-as-a-judge methods~\cite{llm_judge1, llm_judge2}. We summarize each dimension below; full details are in Appendix~\ref{sec:scoring_details}.

\subsubsection{Usability Score}
Usability directly measures downstream task performance. For \textit{entity linking}, it is the Mean Reciprocal Rank (MRR) over $|Q|$ evaluation queries:
\[
S_{\text{usability}} = \frac{1}{|Q|} \sum_{i=1}^{|Q|} \frac{1}{\text{rank}_i} \times 100,
\]
where $\text{rank}_i$ is the rank of the correct field for query $i$. When downstream evaluation is unavailable, the remaining four metrics still provide meaningful assessment.

\subsubsection{Informativeness Score}
Informativeness captures the diversity of information in a description. We split the text into clause components, embed each with a text embedding model, and compute the average pairwise cosine similarity $\bar{s}$:
\[
S_{\mathrm{info}} = 100 \times \,(1-\bar{s}), \quad S_{\mathrm{info}}=0 \text{ if } m<2.
\]

\subsubsection{Clarity Score}
Clarity measures interpretation consistency. Inspired by~\cite{clarifyGPT}, we prompt an LLM to generate $N$ independent interpretations, embed them, and compute mean pairwise similarity:
\[
S_{\text{clarity}} = \frac{2}{N(N-1)} \sum_{i<j} {\rm CosSim}(E_i, E_j) \times 100.
\]

\subsubsection{Conciseness Score}
Conciseness detects redundancy via both syntactic (Jaccard~\cite{jaccard}) and semantic (embedding cosine) similarity between sentence pairs. Let $N_{\text{redundant}}$ be the number of sentences in at least one redundant pair out of $N_{\text{total}}$:
\[
S_{\text{concise}} = 100 \times \bigl(1 - \frac{N_{\text{redundant}}}{N_{\text{total}}}\bigr).
\]

\subsubsection{Readability Score}
We adopt the Flesch--Kincaid Reading Ease score~\cite{flesch_score} as the readability metric.



\begin{table*}[t]
\centering
\small
\caption{Clarification question taxonomy for semantic enrichment.}
\renewcommand{\arraystretch}{1.2}
\setlength{\tabcolsep}{4pt}
\begin{tabular}{@{}p{2.2cm}p{7.2cm}p{5.8cm}@{}}
\toprule
\textbf{Type} & \textbf{Description} & \textbf{Example} \\
\midrule
\textbf{Paraphrase} & When the description is \textcolor{red}{generally ambiguous}, the system can ask the user to paraphrase it in another way. Paraphrasing can cross-validate the meaning and is also easy for users to perform. & 
\textit{User:} ``This is a field about Python.'' \newline
\textit{{\tool}:} ``Can you paraphrase it in another way?'' \\
\midrule
\textbf{Narrow-down} & When there are \textcolor{red}{different meanings} of a certain mention, the system asks a clarification question to narrow down the meaning. & 
\textit{User:} ``Python.'' \newline
\textit{{\tool}:} ``Are you talking about the programming language or the animal (snake)?'' \\
\midrule
\textbf{Attributes} & The mention may not be ambiguous, but there are multiple \textcolor{red}{subtypes/attributes} under it. The system can request more detail and identify which attributes the user cares about. & 
\textit{User:} ``Programming languages.'' \newline
\textit{{\tool}:} ``Do you care more about Python, Java, or other languages? Or is this for all languages?'' \\
\midrule
\textbf{Wh-clarification} & The system asks \textit{Who} (personal context), \textit{When} (temporal context), \textit{Where} (spatial context), or \textit{Why} (purpose context) questions about the description. & 
\textit{User:} ``Laptop.'' \newline
\textit{{\tool}:} ``Is there any purpose, such as gaming, work, or school?'' \\
\midrule
\textbf{Relationship} & There can be a \textcolor{red}{relationship between this field and other fields} in the schema/sandbox, but the user did not mention it. The system identifies a likely candidate and asks if such a relationship exists. & 
\textit{User:} ``This field collects user names who understand Python programming.'' \newline
\textit{{\tool}:} ``Is there a relationship between this field and the other field \textit{expertise\_level}?'' \\
\midrule
\textbf{Logic} & The \textcolor{red}{logic} presented in the description can be interpreted in different ways. & 
\textit{User:} ``Python and Java.'' \newline
\textit{{\tool}:} ``Do you mean `either' or `both' for the two conditions?'' \\
\bottomrule
\end{tabular}
\label{tab:clarification-taxonomy}
\end{table*}

\subsection{Taxonomy-Guided Clarification}
\label{sec:clarification}


To add missing context and resolve ambiguity, we propose a clarification taxonomy to support clarification question generation for field enrichment, based on previous studies~\cite{taxonomy1, taxonomy2, taxonomy3}.

As shown in Table~\ref{tab:clarification-taxonomy}, each type targets a distinct cause of incompleteness or ambiguity:
\textit{Paraphrase} asks the user to restate the description, helping cross-validate and disambiguate meaning.
\textit{Narrow-down} resolves polysemy by restricting the scope of a term with multiple interpretations.
\textit{Attributes} collects finer-grained details when relevant subtypes or attributes are not explicitly specified.
\textit{Wh-clarification} elicits further context by asking who, when, where, or why questions.
\textit{Relationship} uncovers potential connections between the given field and other fields in the schema.
\textit{Logic} clarifies the intended logical relationships between multiple elements or conditions in the description.

Given a field description and its context (e.g., schema information or previously answered questions), {\tool} generates a clarification question in two steps.
First, it prompts an LLM in an ``LLM-as-a-Judge'' role to select the most relevant clarification type from the taxonomy.
Second, it builds a type-specific generation prompt containing the description, context, few-shot examples, and a short style template to produce a concise question along with candidate answers (multiple-choice or open-ended).
This two-step pipeline ensures that each generated question is both consistent in form and targeted at the most pressing ambiguity.

\subsection{Conditional Query Population}
\label{sec:query_population}



Without the additional instruction, {\tool} generates diverse queries related to this field by default

Beyond passively answering clarification questions, {\tool} allows users to proactively provide NL queries or example data that convey the purpose or usage scenario of a data field.
Since manually creating such examples can be demanding, {\tool} introduces a query population feature that automatically generates candidate queries for a given field. Our key insight is that verifying related queries is often easier than refining a textual description, so users only need to validate or refine populated candidates rather than author them from scratch.
The process is iterative: previously verified or user-provided queries become context for subsequent rounds, progressively improving semantic coverage.
Users can optionally steer generation with an NL instruction (e.g., ``focus on time evaluation'' to populate only chronological queries); without one, {\tool} generates diverse queries by default.

\subsection{User Interface}

\begin{figure*}[t]
\centering
\includegraphics[width=\textwidth]{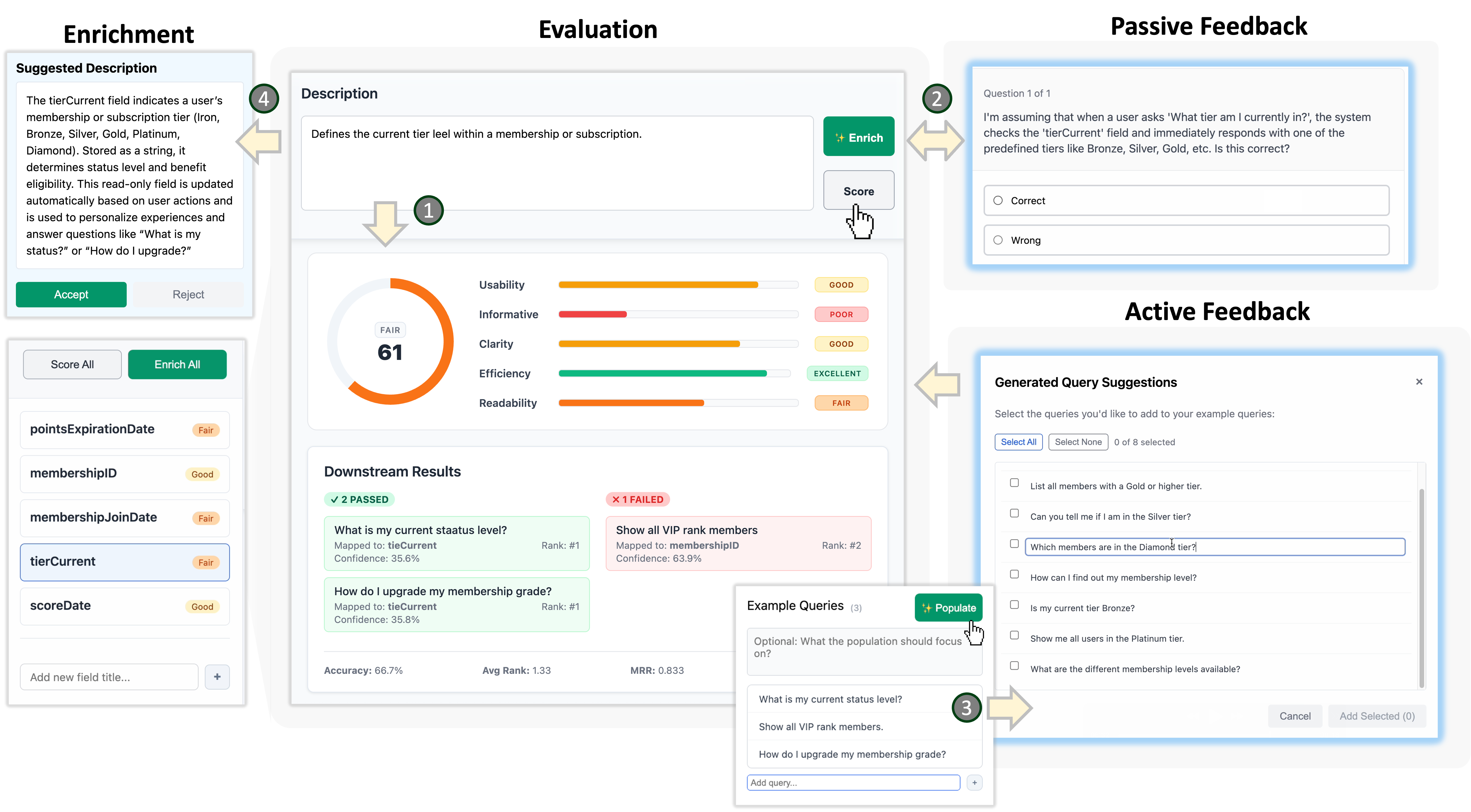}
\caption{UI of {\tool}.
\textbf{(1) Evaluate:} Users click \emph{Score} to obtain a 5-dimensional 0–100 scores and a \emph{Downstream Results} dashboard to indicate the quality of the current description. 
\textbf{(2) Passive feedback:} The UI asks concise clarification questions (e.g., scope, attributes, logic); users answer with one click. 
\textbf{(3) Active feedback:} Users add example NL queries, or press \emph{Populate} to review system-populated queries and select the relevant ones.
\textbf{(4) Enrichment:} The system drafts a \emph{Suggested Description} that the user can Accept/Reject or lightly edit. 
}
\label{fig:ui}
\end{figure*}

Figure~\ref{fig:ui} illustrates the UI of {\tool}, which is organized around the three-stage pipeline described in Section~\ref{sec:method}.
The interface is designed to minimize cognitive load while keeping the domain expert in control at every step.

The left panel~(\circled{1}) displays the current field description and a \emph{Score} button that triggers the scoring system (Section~\ref{sec:scoring}). Results are shown as a radar chart of the five dimension scores, an overall quality score (0--100) with natural language explanations, and a \emph{Downstream Results} dashboard that reveals the field's ranking performance on real evaluation queries.
The middle-left area~(\circled{2}) presents taxonomy-guided clarification questions (Section~\ref{sec:clarification}) as clickable multiple-choice options or short free-text inputs, enabling users to provide feedback with minimal effort.
The middle-right area~(\circled{3}) provides the query population interface (Section~\ref{sec:query_population}), where users can type NL queries directly or press \emph{Populate} to generate a batch of candidates displayed as selectable chips. An optional instruction field lets users steer the generation (e.g., ``focus on time-related queries''), and selected queries accumulate across rounds.
The right panel~(\circled{4}) shows a \emph{Suggested Description} synthesized from all collected feedback, which users can accept, reject, or edit in place. A description history viewer lets users compare successive versions.
The entire workflow is iterative: after accepting or editing a suggestion, users can re-score the updated description and continue refining until the quality meets their expectations.

\section{Evaluation}

\subsection{User Study}

As an interactive tool, we evaluate the usability and effectiveness of {\tool}.
We first conducted a within-subjects user study with 8 participants.
All of them engage with data field descriptions as part of their day-to-day work.
In the study, we compare {\tool} to three baselines, including manually editing the description, selecting enriched description candidates, and using a conversational AI assistant.
Following the user study, we conducted an additional qualitative evaluation in which human reviewers rated the enriched descriptions against the ground truth.
The ratings were conducted with the reviewers unaware of the origin of the descriptions, as descriptions were shuffled across all users and sessions. This approach ensured an unbiased evaluation of whether {\tool} enhances enrichment accuracy.

\subsubsection{User Study Protocol}
Each study began with a two-minute introduction that outlined the motivation and background of the study.
Next, participants were asked to complete a pre-task survey to capture their background information and better understand user needs.
To simulate a scenario that elicits domain experts' knowledge, participants were given three minutes to memorize a printed sheet containing the ground truth descriptions of four data fields, which would later serve as the study tasks.
The sheet was then collected to simulate a scenario in which participants hold knowledge in their minds without documented information.
Following this, participants watched a three-minute tutorial video explaining the study tasks and demonstrating how to use {\tool} and the other tools.
Participants proceeded to complete four task sessions, each lasting three minutes and utilizing a different tool. 
Participants were asked to make their best effort to provide a good description within the allotted time for each session.
Both the order of tools and their assigned tasks were shuffled to negate learning effects.
Finally, after completing all tasks, participants filled out a post-task survey to compare their experiences across the different sessions.

\subsubsection{Comparison Baselines}

To evaluate the effectiveness of {\tool}, we compared it against three baselines that reflect common practices for enriching data field descriptions:

\begin{itemize}
    \item \textbf{Manual Editing.} Participants directly wrote or revised the field description without system support. This baseline simulates the most common practice, where users rely solely on their own domain knowledge.
    
    \item \textbf{Candidate Selection.} Participants were presented with several enriched descriptions generated by LLM prompting. They could select one of the candidates and manually edit it. 
    
    \item \textbf{Conversational AI Assistant.} Participants interacted with a general-purpose conversational AI assistant, ChatGPT. Users iteratively refine the description through natural chatting, a process with which they are already familiar.
\end{itemize}


\subsubsection{Pre-task Survey}



Before starting the main tasks, participants completed a short pre-task survey to capture their prior experience and perceptions of field descriptions. All eight participants reported that they were not satisfied with the quality of the field descriptions that appeared in their work. In terms of usage frequency, six participants indicated that they interact with database fields on a daily basis, while the remaining two reported doing so on a weekly basis.

Participants also identified field description issues in their work. As shown in Table~\ref{tab:field-challenges}, the most frequent issue was that descriptions are too short or missing (100\%). Other common limitations included a lack of concrete examples (71.4\%), missing information about relationships between fields (71.4\%), and insufficient business context or purpose (57.1\%). Ambiguity in language (42.9\%) and outdated or incorrect descriptions (28.6\%) were also noted. 
These findings confirm that field descriptions often fail to meet practitioners’ needs and motivate the design of {\tool} to provide more complete, contextualized, and useful enrichment.

\begin{figure*}[ht]
\centering
\includegraphics[width=0.74\textwidth]{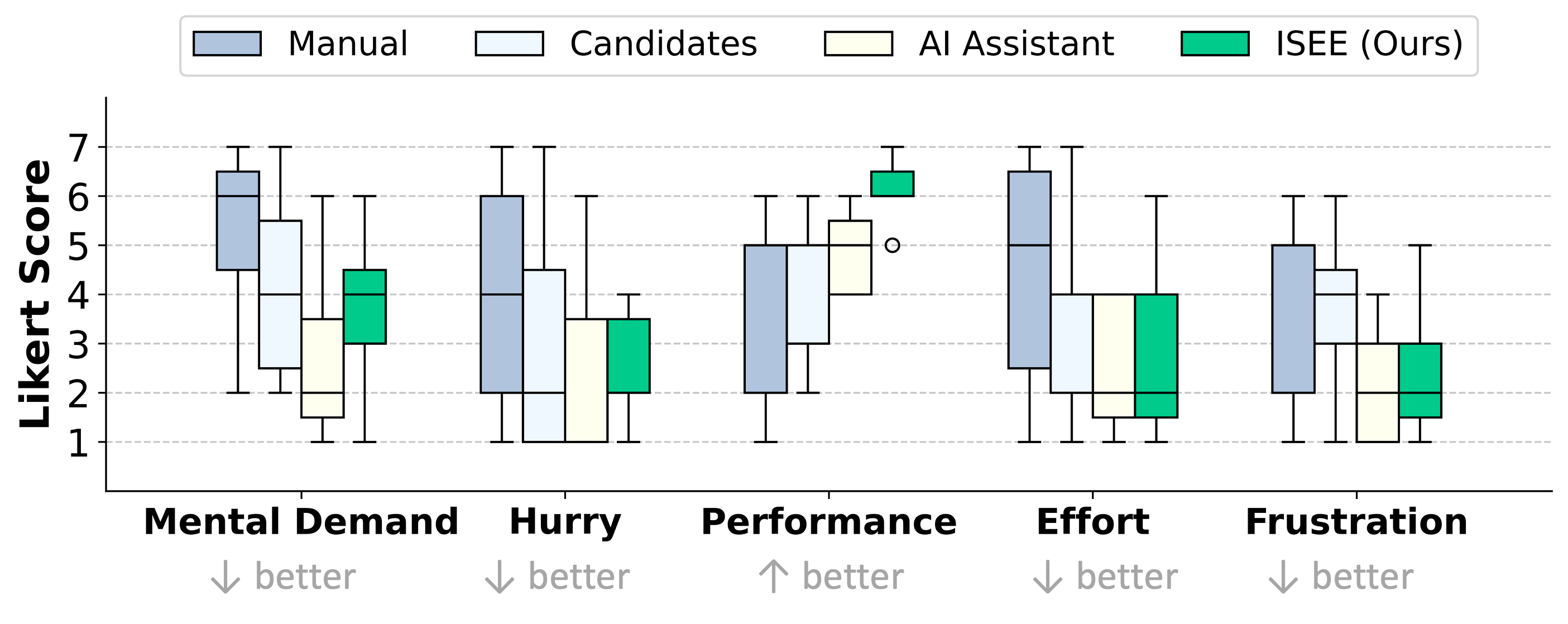}
\caption{NASA Task Load Index Ratings.}
\label{fig:nasa}
\end{figure*}

Furthermore, we asked participants to reflect on the important aspects of field descriptions. 
Results are summarized in Table~\ref{tab:field-importance}. \textit{Field purpose} and \textit{Data type/format} were consistently ranked as the most important content of a description. 

\subsubsection{User Study Results}

We first evaluated {\tool} against three baselines (manual editing, candidate selection, and conversational assistant) using the NASA TLX questionnaire~\cite{nasa_tlx}. As shown in Figure~\ref{fig:nasa}, {\tool} significantly reduced perceived mental demand, effort, and frustration compared to the baselines. The reductions in mental demand and effort were statistically significant ($p<0.05$). Participants reported that enriching field descriptions with {\tool} required less cognitive load and felt smoother than manually editing or relying on a conversational assistant. Importantly, the highest ratings on performance indicate that participants felt more successful in completing the enrichment task when using {\tool}.

To further assess enrichment quality, we conducted a blind review in which independent human raters scored enriched descriptions against ground-truth references without knowing their source (Table~\ref{tab:qualitative}). Descriptions generated with {\tool} received significantly higher accuracy ratings than all baselines, and this difference was statistically significant ($p<0.05$), which is consistent with participants' self-perception.

\begin{table}[t]
\centering
\small
\caption{Participants reported limitations of field descriptions (7 participants responded, multiple choices allowed).}
\renewcommand{\arraystretch}{1.1}
\setlength{\tabcolsep}{4pt}
\begin{tabular}{@{}lc@{}}
\toprule
\textbf{Challenge} & \textbf{Responses (\%)} \\
\midrule
Too short or missing & 7 (100\%) \\
Ambiguous or unclear language & 3 (42.9\%) \\
Outdated or incorrect & 2 (28.6\%) \\
Lacks business context or purpose & 4 (57.1\%) \\
Lacks concrete examples & 5 (71.4\%) \\
Lacks relationship information & 5 (71.4\%) \\
Other & 0 (0\%) \\
\bottomrule
\end{tabular}
\label{tab:field-challenges}
\end{table}

\begin{table}[t]
\centering
\small
\caption{Average participant rankings of contents of field descriptions (lower score = higher importance).}
\renewcommand{\arraystretch}{1.1}
\setlength{\tabcolsep}{4pt}
\begin{tabular}{@{}lc@{}}
\toprule
\textbf{Content} & \textbf{Avg.\ Rank} \\
\midrule
Data Type and Format & 1.57 \\
Field Purpose & 2.00 \\
Usage Example & 2.86 \\
Field Relationship & 3.14 \\
\bottomrule
\end{tabular}
\label{tab:field-importance}
\end{table}

Additionally, in the post-study survey, we gathered additional feedback. The majority of participants rated the quality score as useful or very useful, with 85.8\% assigning a score of 4 or 5 on a 5-point scale. 
We also asked participants to rate the usefulness of three key features of our system on a 1–5 scale (higher is better).
Query population received the highest rating ($M=4.29$), with participants emphasizing that the automatically generated queries were ``\textit{diverse in nature, covering many possible questions a user may have about the field,}'' which ``\textit{made it much easier to write a good description.}'' Description scoring was also rated highly ($M=3.86$), with one participant noting that ``\textit{the scoring of the interactive enrichment system is extremely helpful as it can guide me to improve my description.}'' Finally, clarification questions received a relatively lower rating ($M=2.71$). While generally considered useful, participants mentioned that ``\textit{sometimes [they] felt it less efficient to answer many questions.}'' Taken together, these results suggest that all three features contribute to the effectiveness of semantic enrichment, with query population and description scoring being particularly impactful.

\subsection{Automated User Simulation}

To evaluate {\tool} at scale beyond the user study, we follow prior work~\cite{steps, misp} and design an user simulation experiment where an LLM simulates a domain expert interacting with {\tool}.
We experiment on the BIRD benchmark~\cite{birdbench}, a public text-to-SQL dataset containing real-world databases with column descriptions. We evaluate enrichment quality through two downstream tasks: \emph{entity linking}, where we rank all columns by relevance to a given NL question and calculate Mean Reciprocal Rank (MRR) of the gold columns; and \emph{text-to-SQL}, where we generate SQL using GPT-4o with column descriptions as context and report execution accuracy (EX), which checks whether the predicted SQL produces the same result as the gold query, and Valid Efficiency Score (VES), which combines execution accuracy with query efficiency by penalizing slow queries.

\begin{table}[t]
\centering
\small
\caption{Average reviewer ratings of enriched descriptions (1 = lowest quality, 7 = highest quality).}
\renewcommand{\arraystretch}{1.1}
\setlength{\tabcolsep}{4pt}
\begin{tabular}{@{}lc@{}}
\toprule
\textbf{Method} & \textbf{Rating (1--7)} \\
\midrule
Manual Editing & 3.2 \\
Candidate Selection & 4.0 \\
Conversational AI Assistant & 4.5 \\
\midrule
{\tool} (Ours) & \textbf{6.2} \\
\bottomrule
\end{tabular}
\label{tab:qualitative}
\end{table}

We construct a \emph{reference description} for each column by combining its existing BIRD description, schema (table, columns, foreign keys), sampled database records, and evidence from gold SQL queries and their associated NL questions. This reference represents the domain knowledge an expert would hold and is used only to drive the simulated user's responses, which is \emph{never} provided to {\tool}.
The simulated user (a persona prompt containing the reference context) can only interact through {\tool}'s UI features: answering clarification questions and accepting or rejecting populated queries. It cannot directly write or paste the description.

We run the simulation on the entire BIRD dev set (1534 queries, 11 databases) and compare among three conditions: (1) \emph{Original}: the existing BIRD description; (2) \emph{Auto-enrich}: a single-pass LLM baseline that improves the description using only schema context, without interactive feedback or the reference; and (3) \emph{{\tool} (simulated)}: enrichment through the simulated user interacting with {\tool}.
As shown in Table~\ref{tab:simulation}, {\tool} with simulated users significantly outperforms both baselines. Compared to Auto-enrich, {\tool} improves entity-linking MRR by 9.47 and text-to-SQL EX by 6.45, demonstrating that interactive clarification and evaluation can elicit domain knowledge that single generation cannot cover. Regarding token cost per column, Auto-enrich requires on average 150 output tokens, while {\tool} consumes on average 380 output tokens.

\begin{table}[t]
\centering
\small
\caption{User simulation results on full BIRD.}
\renewcommand{\arraystretch}{1.1}
\setlength{\tabcolsep}{4pt}
\begin{tabular}{@{}lccc@{}}
\toprule
& \textbf{Entity Linking} & \multicolumn{2}{c}{\textbf{Text-to-SQL}} \\
\cmidrule(lr){2-2} \cmidrule(lr){3-4}
\textbf{Method} & MRR & EX & VES \\
\midrule
Original & 42.31 & 52.22 & 56.99 \\
Auto-enrich & 55.25 & 54.95 & 60.18 \\
\midrule
{\tool} (simulated) & 64.72 & 61.40 & 66.54 \\
\bottomrule
\end{tabular}
\label{tab:simulation}
\end{table}

\subsection{Quantitative Evaluation}

To complement the user simulation experiment, we further conducted a quantitative experiment where one author interactively enriches descriptions using {\tool}. We randomly sampled 100 NL questions across 18 databases covering 135 columns from BIRD. The experimenter worked on each column description within a 3-minute time limit. Since the experimenter is fully familiar with the system, these results can be interpreted as an upper bound of user performance with {\tool}.

Table~\ref{tab:quant-results} reports the results. Enrichment with {\tool} improves entity-linking MRR by 21.16, text-to-SQL execution accuracy by 9.00, and VES by 5.88 over the original descriptions. {\tool} outperforms Auto-enrich across all metrics, confirming that interactive enrichment yields higher-quality descriptions than automated generation.

\begin{table}[t]
\centering
\small
\caption{Quantitative evaluation on sampled BIRD.}
\renewcommand{\arraystretch}{1.1}
\setlength{\tabcolsep}{4pt}
\begin{tabular}{@{}lccc@{}}
\toprule
& \textbf{Entity Linking} & \multicolumn{2}{c}{\textbf{Text-to-SQL}} \\
\cmidrule(lr){2-2} \cmidrule(lr){3-4}
\textbf{Method} & MRR & EX & VES \\
\midrule
Original & 45.83 & 55.00 & 58.77 \\
Auto-enrich & 55.01 & 57.00 & 63.80 \\
\midrule
{\tool} & 66.99 & 64.00 & 64.65 \\
\bottomrule
\end{tabular}
\label{tab:quant-results}
\end{table}

\subsection{Case Study}
We conduct a case study to understand how enrichment with {\tool} affects downstream task performance in a realistic setting.
In our study, one participant was asked to enrich the field \texttt{purchases.value}, whose original description was simply \textit{``Value.''} The database also contains fields such as \texttt{order.priceTotal} (\textit{``Total price of the order.''}) and \texttt{purchases.id} (\textit{``Purchase identifier.''}). We examine how a user query, \textit{``What is the total revenue from purchases last quarter?''}, is handled before and after enrichment.

\textbf{Before enrichment.}
The entity-linking system cannot determine what \textit{``value''} refers to. It ranks \texttt{order.priceTotal} first (since its description explicitly mentions \textit{``price''}) and \texttt{purchases.value} third. Consequently, a text-to-SQL system generates an incorrect query: \texttt{\textcolor[RGB]{172,41,0}{SELECT SUM}(order.priceTotal) \textcolor[RGB]{172,41,0}{FROM} order \textcolor[RGB]{172,41,0}{WHERE} order\_date >= '2025-10-01'}. This returns the sum of all order totals, not the transaction revenue.

\textbf{Enrichment with {\tool}.}
Through clarification, the participant specifies that the field refers to a \textit{``monetary amount''} at the \textit{``per-transaction''} level. Through query population, the participant verifies relevant queries such as \textit{``average purchase amount per customer.''} The description becomes: \textit{``The monetary amount (in base currency) of each completed purchase transaction. Used for revenue aggregation. Distinct from order-level totals such as order.priceTotal.''} The score rises from 18 to 82.

\textbf{After enrichment.}
With the enriched description explicitly mentioning \textit{``monetary amount''} and \textit{``revenue aggregation,''} entity linking now correctly ranks \texttt{purchases.value} first, as its description directly matches the user's intent. The text-to-SQL system accordingly generates the correct query: \texttt{\textcolor[RGB]{0,120,60}{SELECT SUM}(purchases.value) \textcolor[RGB]{0,120,60}{FROM} purchases \textcolor[RGB]{0,120,60}{WHERE} purchase\_date >= '2025-10-01'}, which correctly computes revenue from purchase transactions rather than order totals. This example illustrates how a single enrichment session with {\tool} can resolve ambiguity that otherwise propagates through the downstream pipeline.

\section{Conclusion}
We present {\tool}, an interactive semantic enrichment system that automatically scores field descriptions, and improve it through query population and clarification questions. 
A user study shows that {\tool} significantly reduces cognitive load while improving performance.
A user simulation experiment and a quantitative evaluation shows that data descriptions enriched by {\tool} can significantly benefit downstream tasks.
This work demonstrates that interactive enrichment is a promising approach for capturing missing domain knowledge and benefiting data-related applications.

\section*{Limitations}

Like any interactive system that relies on human expertise, {\tool} assumes that the user possesses domain knowledge about the fields to enrich.
Regarding practical use, while users are domain experts who defined or work with these fields daily, their knowledge may occasionally be incomplete or outdated.
Future work could incorporate confidence estimation or cross-validation mechanisms to further safeguard enrichment quality.

On the evaluation side, we designed three complementary studies to balance ecological validity with scale. Our user study recruited eight professional industry engineers and researchers with hands-on data management experience, ensuring high-quality feedback, though the sample size is naturally constrained by the cost of expert recruitment. To scale beyond this, we introduced an automated user simulation over the entire BIRD dev set; while this enables comprehensive coverage, simulated interactions inevitably simplify real human behavior such as partial or ambiguous responses. The quantitative evaluation, where the first author manually enriched sampled databases, bridges these two settings with controlled but smaller-scale evidence. We believe these three evaluations together offer a thorough and balanced assessment of {\tool}.



\bibliography{custom}

@ARTICLE{entity_linking1,
  author={Shen, Wei and Wang, Jianyong and Han, Jiawei},
  journal={IEEE Transactions on Knowledge and Data Engineering}, 
  title={Entity Linking with a Knowledge Base: Issues, Techniques, and Solutions}, 
  year={2015},
  volume={27},
  number={2},
  pages={443-460},
  doi={10.1109/TKDE.2014.2327028},
  ISSN={1558-2191},
  month={Feb},
}

@misc{entity_linking2,
      title={End-to-End Neural Entity Linking}, 
      author={Nikolaos Kolitsas and Octavian-Eugen Ganea and Thomas Hofmann},
      year={2018},
      eprint={1808.07699},
      archivePrefix={arXiv},
      primaryClass={cs.CL},
      url={https://arxiv.org/abs/1808.07699}, 
}

@misc{sqlucid,
      title={SQLucid: Grounding Natural Language Database Queries with Interactive Explanations}, 
      author={Yuan Tian and Jonathan K. Kummerfeld and Toby Jia-Jun Li and Tianyi Zhang},
      year={2024},
      eprint={2409.06178},
      archivePrefix={arXiv},
      primaryClass={cs.HC},
      url={https://arxiv.org/abs/2409.06178}, 
}

@inproceedings{sqlsynth,
author = {Tian, Yuan and Lee, Daniel and Wu, Fei and Mai, Tung and Qian, Kun and Sahai, Siddhartha and Zhang, Tianyi and Li, Yunyao},
title = {Text-to-SQL Domain Adaptation via Human-LLM Collaborative Data Annotation},
year = {2025},
isbn = {9798400713064},
publisher = {Association for Computing Machinery},
address = {New York, NY, USA},
url = {https://doi.org/10.1145/3708359.3712083},
doi = {10.1145/3708359.3712083},
booktitle = {Proceedings of the 30th International Conference on Intelligent User Interfaces},
pages = {1398–1425},
numpages = {28},
location = {
},
series = {IUI '25}
}

@inproceedings{steps,
    title = "Interactive Text-to-{SQL} Generation via Editable Step-by-Step Explanations",
    author = "Tian, Yuan  and
      Zhang, Zheng  and
      Ning, Zheng  and
      Li, Toby Jia-Jun  and
      Kummerfeld, Jonathan K.  and
      Zhang, Tianyi",
    editor = "Bouamor, Houda  and
      Pino, Juan  and
      Bali, Kalika",
    booktitle = "Proceedings of the 2023 Conference on Empirical Methods in Natural Language Processing",
    month = dec,
    year = "2023",
    address = "Singapore",
    publisher = "Association for Computational Linguistics",
    url = "https://aclanthology.org/2023.emnlp-main.1004/",
    doi = "10.18653/v1/2023.emnlp-main.1004",
    pages = "16149--16166"
}

@article{insight,
author = {Ning, Zheng and Tian, Yuan and Zhang, Zheng and Zhang, Tianyi and Li, Toby Jia-Jun},
title = {Insights into Natural Language Database Query Errors: from Attention Misalignment to User Handling Strategies},
year = {2024},
issue_date = {December 2024},
publisher = {Association for Computing Machinery},
address = {New York, NY, USA},
volume = {14},
number = {4},
issn = {2160-6455},
url = {https://doi.org/10.1145/3650114},
doi = {10.1145/3650114},
journal = {ACM Trans. Interact. Intell. Syst.},
month = dec,
articleno = {25},
numpages = {32}
}

@inproceedings{empirical,
author = {Ning, Zheng and Zhang, Zheng and Sun, Tianyi and Tian, Yuan and Zhang, Tianyi and Li, Toby Jia-Jun},
title = {An Empirical Study of Model Errors and User Error Discovery and Repair Strategies in Natural Language Database Queries},
year = {2023},
isbn = {9798400701061},
publisher = {Association for Computing Machinery},
address = {New York, NY, USA},
url = {https://doi.org/10.1145/3581641.3584067},
doi = {10.1145/3581641.3584067},
booktitle = {Proceedings of the 28th International Conference on Intelligent User Interfaces},
pages = {633–649},
numpages = {17},
location = {Sydney, NSW, Australia},
series = {IUI '23}
}

@article{clarifyGPT,
author = {Mu, Fangwen and Shi, Lin and Wang, Song and Yu, Zhuohao and Zhang, Binquan and Wang, ChenXue and Liu, Shichao and Wang, Qing},
title = {ClarifyGPT: A Framework for Enhancing LLM-Based Code Generation via Requirements Clarification},
year = {2024},
issue_date = {July 2024},
publisher = {Association for Computing Machinery},
address = {New York, NY, USA},
volume = {1},
number = {FSE},
url = {https://doi.org/10.1145/3660810},
doi = {10.1145/3660810},
journal = {Proc. ACM Softw. Eng.},
month = jul,
articleno = {103},
numpages = {23}
}

@misc{spa,
      title={Selective Prompt Anchoring for Code Generation}, 
      author={Yuan Tian and Tianyi Zhang},
      year={2025},
      eprint={2408.09121},
      archivePrefix={arXiv},
      primaryClass={cs.LG},
      url={https://arxiv.org/abs/2408.09121}, 
}

@book{flesch_score,
  author    = {Rudolf Franz Flesch},
  title     = {Marks of Readable Style: A Study in Adult Education},
  year      = {1943},
  publisher = {Teachers College, Columbia University},
  language  = {English},
  url       = {https://nla.gov.au/nla.cat-vn916503},
  note      = {Accessed: 30 January 2025, via National Library of Australia}
}

@phdthesis{taxonomy1,
  author       = {Purver, Matthew R.\,J.},
  title        = {The Theory and Use of Clarification Requests in Dialogue},
  school       = {King’s College London},
  year         = {2004},
  note         = {PhD thesis},
}

@inproceedings{taxonomy2,
author = {Aliannejadi, Mohammad and Zamani, Hamed and Crestani, Fabio and Croft, W. Bruce},
title = {Asking Clarifying Questions in Open-Domain Information-Seeking Conversations},
year = {2019},
isbn = {9781450361729},
publisher = {Association for Computing Machinery},
address = {New York, NY, USA},
url = {https://doi.org/10.1145/3331184.3331265},
doi = {10.1145/3331184.3331265},
booktitle = {Proceedings of the 42nd International ACM SIGIR Conference on Research and Development in Information Retrieval},
pages = {475–484},
numpages = {10},
location = {Paris, France},
series = {SIGIR'19}
}

@inproceedings{taxonomy3,
author = {Zamani, Hamed and Dumais, Susan and Craswell, Nick and Bennett, Paul and Lueck, Gord},
title = {Generating Clarifying Questions for Information Retrieval},
year = {2020},
isbn = {9781450370233},
publisher = {Association for Computing Machinery},
address = {New York, NY, USA},
url = {https://doi.org/10.1145/3366423.3380126},
doi = {10.1145/3366423.3380126},
booktitle = {Proceedings of The Web Conference 2020},
pages = {418–428},
numpages = {11},
location = {Taipei, Taiwan},
series = {WWW '20}
}

@misc{gpt4o,
      title={GPT-4o System Card}, 
      author={OpenAI},
      year={2024},
      eprint={2410.21276},
      archivePrefix={arXiv},
      primaryClass={cs.CL},
      url={https://arxiv.org/abs/2410.21276}, 
}

@article{semantic_enrichment1,
  title={Semantic enrichment for building information modeling},
  author={Belsky, Michael and Sacks, Rafael and Brilakis, Ioannis},
  journal={Computer-Aided Civil and Infrastructure Engineering},
  volume={31},
  number={4},
  pages={261--274},
  year={2016},
  publisher={Wiley Online Library}
}

@misc{semantic_enrichment2,
      title={Columbo: Expanding Abbreviated Column Names for Tabular Data Using Large Language Models}, 
      author={Ting Cai and Stephen Sheen and AnHai Doan},
      year={2025},
      eprint={2508.09403},
      archivePrefix={arXiv},
      primaryClass={cs.CL},
      url={https://arxiv.org/abs/2508.09403}, 
}

@article{semantic_enrichment3,
title = {Semantic enrichment of building and city information models: A ten-year review},
journal = {Advanced Engineering Informatics},
volume = {47},
pages = {101245},
year = {2021},
issn = {1474-0346},
doi = {https://doi.org/10.1016/j.aei.2020.101245},
url = {https://www.sciencedirect.com/science/article/pii/S1474034620302147},
author = {Fan Xue and Liupengfei Wu and Weisheng Lu}
}

@article{jaccard,
  title={Distance between sets},
  author={Levandowsky, Michael and Winter, David},
  journal={Nature},
  volume={234},
  number={5323},
  pages={34--35},
  year={1971},
  publisher={Nature Publishing Group},
  doi={10.1038/234034a0}
}

@incollection{nasa_tlx,
title = {Development of NASA-TLX (Task Load Index): Results of Empirical and Theoretical Research},
editor = {Peter A. Hancock and Najmedin Meshkati},
series = {Advances in Psychology},
publisher = {North-Holland},
volume = {52},
pages = {139-183},
year = {1988},
booktitle = {Human Mental Workload},
issn = {0166-4115},
doi = {https://doi.org/10.1016/S0166-4115(08)62386-9},
url = {https://www.sciencedirect.com/science/article/pii/S0166411508623869},
author = {Sandra G. Hart and Lowell E. Staveland}
}

@misc{llm_score1,
      title={LLM as a Scorer: The Impact of Output Order on Dialogue Evaluation}, 
      author={Yi-Pei Chen and KuanChao Chu and Hideki Nakayama},
      year={2024},
      eprint={2406.02863},
      archivePrefix={arXiv},
      primaryClass={cs.CL},
      url={https://arxiv.org/abs/2406.02863}, 
}

@misc{llm_score2,
      title={Beyond Yes and No: Improving Zero-Shot LLM Rankers via Scoring Fine-Grained Relevance Labels}, 
      author={Honglei Zhuang and Zhen Qin and Kai Hui and Junru Wu and Le Yan and Xuanhui Wang and Michael Bendersky},
      year={2024},
      eprint={2310.14122},
      archivePrefix={arXiv},
      primaryClass={cs.IR},
      url={https://arxiv.org/abs/2310.14122}, 
}

@misc{llm_score3,
      title={Calibrating LLM-Based Evaluator}, 
      author={Yuxuan Liu and Tianchi Yang and Shaohan Huang and Zihan Zhang and Haizhen Huang and Furu Wei and Weiwei Deng and Feng Sun and Qi Zhang},
      year={2023},
      eprint={2309.13308},
      archivePrefix={arXiv},
      primaryClass={cs.CL},
      url={https://arxiv.org/abs/2309.13308}, 
}

@misc{llm_judge1,
      title={A Survey on LLM-as-a-Judge}, 
      author={Jiawei Gu and Xuhui Jiang and Zhichao Shi and Hexiang Tan and Xuehao Zhai and Chengjin Xu and Wei Li and Yinghan Shen and Shengjie Ma and Honghao Liu and Saizhuo Wang and Kun Zhang and Yuanzhuo Wang and Wen Gao and Lionel Ni and Jian Guo},
      year={2025},
      eprint={2411.15594},
      archivePrefix={arXiv},
      primaryClass={cs.CL},
      url={https://arxiv.org/abs/2411.15594}, 
}

@misc{llm_judge2,
      title={From Generation to Judgment: Opportunities and Challenges of LLM-as-a-judge}, 
      author={Dawei Li and Bohan Jiang and Liangjie Huang and Alimohammad Beigi and Chengshuai Zhao and Zhen Tan and Amrita Bhattacharjee and Yuxuan Jiang and Canyu Chen and Tianhao Wu and Kai Shu and Lu Cheng and Huan Liu},
      year={2025},
      eprint={2411.16594},
      archivePrefix={arXiv},
      primaryClass={cs.AI},
      url={https://arxiv.org/abs/2411.16594}, 
}

@inproceedings{hai1,
author = {Amershi, Saleema and Weld, Dan and Vorvoreanu, Mihaela and Fourney, Adam and Nushi, Besmira and Collisson, Penny and Suh, Jina and Iqbal, Shamsi and Bennett, Paul N. and Inkpen, Kori and Teevan, Jaime and Kikin-Gil, Ruth and Horvitz, Eric},
title = {Guidelines for Human-AI Interaction},
year = {2019},
isbn = {9781450359702},
publisher = {Association for Computing Machinery},
address = {New York, NY, USA},
url = {https://doi.org/10.1145/3290605.3300233},
doi = {10.1145/3290605.3300233},
booktitle = {Proceedings of the 2019 CHI Conference on Human Factors in Computing Systems},
pages = {1–13},
numpages = {13},
location = {Glasgow, Scotland Uk},
series = {CHI '19}
}

@inproceedings{hai2,
author = {Horvitz, Eric},
title = {Principles of mixed-initiative user interfaces},
year = {1999},
isbn = {0201485591},
publisher = {Association for Computing Machinery},
address = {New York, NY, USA},
url = {https://doi.org/10.1145/302979.303030},
doi = {10.1145/302979.303030},
booktitle = {Proceedings of the SIGCHI Conference on Human Factors in Computing Systems},
pages = {159–166},
numpages = {8},
location = {Pittsburgh, Pennsylvania, USA},
series = {CHI '99}
}

@inproceedings{10.1145/3366423.3380126,
author = {Zamani, Hamed and Dumais, Susan and Craswell, Nick and Bennett, Paul and Lueck, Gord},
title = {Generating Clarifying Questions for Information Retrieval},
year = {2020},
isbn = {9781450370233},
publisher = {Association for Computing Machinery},
address = {New York, NY, USA},
url = {https://doi.org/10.1145/3366423.3380126},
doi = {10.1145/3366423.3380126},
booktitle = {Proceedings of The Web Conference 2020},
pages = {418–428},
numpages = {11},
location = {Taipei, Taiwan},
series = {WWW '20}
}

@inproceedings{Aliannejadi_2019, 
    series={SIGIR ’19},
   title={Asking Clarifying Questions in Open-Domain Information-Seeking Conversations},
   url={http://dx.doi.org/10.1145/3331184.3331265},
   DOI={10.1145/3331184.3331265},
   booktitle={Proceedings of the 42nd International ACM SIGIR Conference on Research and Development in Information Retrieval},
   publisher={ACM},
   author={Aliannejadi, Mohammad and Zamani, Hamed and Crestani, Fabio and Croft, W. Bruce},
   year={2019},
   month=jul, pages={475–484},
   collection={SIGIR ’19} }

@article{birdbench,
  title={Can llm already serve as a database interface? a big bench for large-scale database grounded text-to-sqls},
  author={Li, Jinyang and Hui, Binyuan and Qu, Ge and Yang, Jiaxi and Li, Binhua and Li, Bowen and Wang, Bailin and Qin, Bowen and Geng, Ruiying and Huo, Nan and others},
  journal={Advances in Neural Information Processing Systems},
  volume={36},
  year={2024}
}

@misc{singh2025leveragingretrievalaugmentedgenerative,
      title={Leveraging Retrieval Augmented Generative LLMs For Automated Metadata Description Generation to Enhance Data Catalogs}, 
      author={Mayank Singh and Abhijeet Kumar and Sasidhar Donaparthi and Gayatri Karambelkar},
      year={2025},
      eprint={2503.09003},
      archivePrefix={arXiv},
      primaryClass={cs.IR},
      url={https://arxiv.org/abs/2503.09003}, 
}

@misc{gao2025automaticdatabasedescriptiongeneration,
      title={Automatic database description generation for Text-to-SQL}, 
      author={Yingqi Gao and Zhiling Luo},
      year={2025},
      eprint={2502.20657},
      archivePrefix={arXiv},
      primaryClass={cs.AI},
      url={https://arxiv.org/abs/2502.20657}, 
}

@article{zhang2025autoddg,
  title={AutoDDG: Automated Dataset Description Generation using Large Language Models},
  author={Zhang, Haoxiang and Liu, Yurong and Santos, A{\'e}cio and Freire, Juliana and others},
  journal={arXiv preprint arXiv:2502.01050},
  year={2025}
}

@article{wretblad2024synthetic,
  title={Synthetic SQL Column Descriptions and Their Impact on Text-to-SQL Performance},
  author={Wretblad, Niklas and Holmstr{\"o}m, Oskar and Larsson, Erik and Wiks{\"a}ter, Axel and S{\"o}derlund, Oscar and {\"O}hman, Hjalmar and Pont{\'e}n, Ture and Forsberg, Martin and S{\"o}rme, Martin and Heintz, Fredrik},
  journal={arXiv preprint arXiv:2408.04691},
  year={2024}
}

@inproceedings{tabmeta,
  title={TabMeta: Table Metadata Generation with {LLM}-Curated Dataset and {LLM}-Judges},
  author={Anonymous},
  booktitle={Submitted to ACL Rolling Review - June 2024},
  year={2024},
  url={https://openreview.net/forum?id=NXYVm3AjG2},
  note={under review}
}

@inproceedings{misp,
    title = "Model-based Interactive Semantic Parsing: A Unified Framework and A Text-to-{SQL} Case Study",
    author = "Yao, Ziyu  and
      Su, Yu  and
      Sun, Huan  and
      Yih, Wen-tau",
    editor = "Inui, Kentaro  and
      Jiang, Jing  and
      Ng, Vincent  and
      Wan, Xiaojun",
    booktitle = "Proceedings of the 2019 Conference on Empirical Methods in Natural Language Processing and the 9th International Joint Conference on Natural Language Processing (EMNLP-IJCNLP)",
    month = nov,
    year = "2019",
    address = "Hong Kong, China",
    publisher = "Association for Computational Linguistics",
    url = "https://aclanthology.org/D19-1547/",
    doi = "10.18653/v1/D19-1547",
    pages = "5447--5458"
}

@misc{pv_sql,
      title={PV-SQL: Synergizing Database Probing and Rule-based Verification for Text-to-SQL Agents}, 
      author={Yuan Tian and Tianyi Zhang},
      year={2026},
      eprint={2604.17653},
      archivePrefix={arXiv},
      primaryClass={cs.AI},
      url={https://arxiv.org/abs/2604.17653}, 
}

@article{evo_schema,
author = {Zhang, Tianshu and Qian, Kun and Sahai, Siddhartha and Tian, Yuan and Garg, Shaddy and Sun, Huan and Li, Yunyao},
title = {Evoschema: Towards Text-to-SQL Robustness against Schema Evolution},
year = {2025},
issue_date = {June 2025},
publisher = {VLDB Endowment},
volume = {18},
number = {10},
issn = {2150-8097},
url = {https://doi.org/10.14778/3748191.3748222},
doi = {10.14778/3748191.3748222},
journal = {Proc. VLDB Endow.},
month = jun,
pages = {3655–3668},
numpages = {14}
}

@misc{all_fem,
      title={ALL-FEM: Agentic Large Language models Fine-tuned for Finite Element Methods}, 
      author={Rushikesh Deotale and Adithya Srinivasan and Yuan Tian and Tianyi Zhang and Pavlos Vlachos and Hector Gomez},
      year={2026},
      eprint={2603.21011},
      archivePrefix={arXiv},
      primaryClass={cs.CE},
      url={https://arxiv.org/abs/2603.21011}, 
}

@misc{worker_agent,
      title={Supporting Construction Worker Well-Being with a Multi-Agent Conversational AI System}, 
      author={Fan Yang and Yuan Tian and Jiansong Zhang},
      year={2025},
      eprint={2506.07997},
      archivePrefix={arXiv},
      primaryClass={cs.HC},
      url={https://arxiv.org/abs/2506.07997}, 
}

@misc{ood,
      title={Generalized Out-of-Distribution Detection: A Survey}, 
      author={Jingkang Yang and Kaiyang Zhou and Yixuan Li and Ziwei Liu},
      year={2024},
      eprint={2110.11334},
      archivePrefix={arXiv},
      primaryClass={cs.CV},
      url={https://arxiv.org/abs/2110.11334}, 
}

\appendix

\section{Scoring System Details}
\label{sec:scoring_details}

This appendix provides full descriptions of the five scoring metrics used in {\tool}.

\subsection{Usability Score}
The usability score measures how well a field description supports downstream tasks. We consider this the most important metric, as it directly reflects the practical utility of the description in real-world applications. For the \textit{entity linking} task, the score is based on the Mean Reciprocal Rank (MRR), which captures both top-rank accuracy and fine-grained ranking performance. Given a set of $|Q|$ evaluation queries:
\[
S_{\text{usability}} = \frac{1}{|Q|} \sum_{i=1}^{|Q|} \frac{1}{\text{rank}_i} \times 100
\]
where $\text{rank}_i$ is the position of the correct field in the ranked list of retrieval results for query $i$. Higher scores indicate that the description enables more accurate and precise retrieval. As more downstream tasks are incorporated, this score becomes increasingly representative. When downstream evaluation is unavailable (e.g., due to privacy or insufficient data), the remaining four metrics still provide meaningful assessment.

\subsection{Informativeness Score}
Informativeness captures the richness and diversity of the information conveyed in the description. A more informative description provides multiple, non-overlapping aspects of meaning, offering more context and greater semantics.

We split the text into clause components by periods and commas. Each component is embedded with OpenAI's \texttt{text-embedding-3-small} and L2-normalized. Let $m$ be the number of components and $f(\cdot)$ the embedding function. We compute the average pairwise cosine similarity:
\[
\bar{s}=\frac{2}{m(m-1)}\sum_{i<j}\cos\!\big(f(c_i),\,f(c_j)\big),
\]
and define the informativeness score:
\[
S_{\mathrm{info}} = 100\,(1-\bar{s}), \quad \text{with } S_{\mathrm{info}}=0 \text{ if } m<2.
\]
The score is clipped to $[0, 100]$ to stabilize it and mitigate the impact of outliers.

\subsection{Clarity Score}
Clarity measures the extent to which a description can be interpreted in different ways. A high clarity score indicates that different readers or models are likely to interpret the text consistently.

Inspired by prior work~\cite{clarifyGPT} on ambiguity detection for code generation, we prompt an LLM (GPT-4o) to generate $N$ independent interpretations of the input description, then encode each into a semantic embedding. We compute the mean pairwise cosine similarity:
\[
S_{\text{clarity}} = \frac{2}{N(N-1)} \sum_{i<j} {\rm CosSim}(E_i, E_j) \times 100
\]
Lower agreement indicates higher ambiguity, resulting in a lower clarity score.

\subsection{Conciseness Score}
Conciseness measures whether a description conveys its intended meaning without unnecessary redundancy. Redundancy can distract LLMs~\cite{spa}, thereby reducing downstream performance. We quantify redundancy using both syntactic and semantic similarity:
\begin{itemize}
    \item \textbf{Syntactic redundancy}: Pairwise Jaccard similarity~\cite{jaccard} between word sets of two sentences: $J(A, B) = |A \cap B| / |A \cup B|$. Pairs exceeding a threshold are flagged as redundant.
    \item \textbf{Semantic redundancy}: Cosine similarity between sentence embeddings. High similarity indicates semantically equivalent sentences.
\end{itemize}
The final redundancy set is the union of pairs flagged by either method. Let $N_{\text{total}}$ be the total number of sentences and $N_{\text{redundant}}$ the number in at least one redundant pair:
\[
S_{\text{concise}} = 100 \times \bigl(1 - \frac{N_{\text{redundant}}}{N_{\text{total}}}\bigr)
\]

\subsection{Readability Score}
Readability evaluates whether the structure of a description is easy to comprehend. Although originally designed for humans, poor readability indicates that the description is poorly organized and may also hinder LLM understanding. We use the Flesch--Kincaid Reading Ease score~\cite{flesch_score} directly as our readability metric.

\subsubsection{Overall Score}
The overall score is a weighted sum of the five metrics:
\[
S_{\text{total}} = \sum_{m=1}^{5} w_m \cdot S_m
\]
The weights are hyperparameters that can be adjusted to match different enterprise priorities or domain requirements. In our setting, we use weights 0.50, 0.20, 0.15, 0.10, and 0.05 for usability, informativeness, clarity, conciseness, and readability, respectively. These values were calibrated in consultation with multiple engineers in our working environment, with usability weighted highest as it directly reflects downstream task performance. The primary purpose of the overall score is to provide a quick signal for identifying low-quality field descriptions that need attention, rather than serving as a precise ranking metric; the exact weight values are therefore less critical than their relative ordering.

\end{document}